\documentclass[10pt,twocolumn,letterpaper]{article}

\usepackage{cvpr}              %

\usepackage{colortbl}
\usepackage{xcolor}
\definecolor{low}{rgb}{1,0.8,0.8}   %
\definecolor{mid}{rgb}{1,1,0.6}     %
\definecolor{high}{rgb}{0.8,1,0.8}  %

\definecolor{cvprblue}{rgb}{0.21,0.49,0.74}
\usepackage[pagebackref,breaklinks,colorlinks,allcolors=cvprblue]{hyperref}
\usepackage{tikz}
\usepackage{algorithm2e}

\def\paperID{*****} %
\def\confName{CVPR}
\def\confYear{2025}

\title{Challenges in 3D Data Synthesis for Training Neural Networks on Topological Features}

\author{Dylan Peek\textsuperscript{[0000-0003-3568-7853]}\\
School of Information and Physical Sciences\\
The University of Newcastle, Australia\\
{\tt\small dylan.peek@uon.edu.au}
\and
Matthew P. Skerritt\textsuperscript{[0000-0003-2211-7616]}\\
Dept. of Mathematical and Geospatial Sciences \\
RMIT University, Australia\\
{\tt\small matt.skerritt@rmit.edu.au}
\and
Siddharth Pritam\textsuperscript{[0000-0001-5673-0406]}\\
Computer Science Group\\
Chennai Mathematical Institute, India\\
{\tt\small spritam@cmi.ac.in}
\and
Stephan Chalup\textsuperscript{[0000-0002-7886-3653]}\\
School of Information and Physical Sciences\\
The University of Newcastle, Australia\\
{\tt\small stephan.chalup@newcastle.edu.au}
}

\begin{document}
\maketitle
\begin{abstract}
Topological Data Analysis (TDA) involves techniques of analyzing the underlying structure and connectivity of data. However, traditional methods like persistent homology can be computationally demanding, motivating the development of neural network-based estimators capable of reducing computational overhead and inference time. A key barrier to advancing these methods is the lack of labeled 3D data with class distributions and diversity tailored specifically for supervised learning in TDA tasks. To address this, we introduce a novel approach for systematically generating labeled 3D datasets using the Repulsive Surface algorithm, allowing control over topological invariants, such as hole count. The resulting dataset offers varied geometry with topological labeling, making it suitable for training and benchmarking neural network estimators. This paper uses a synthetic 3D dataset to train a genus estimator network, created using a 3D convolutional transformer architecture. An observed decrease in accuracy as deformations increase highlights the role of not just topological complexity, but also geometric complexity, when training generalized estimators. This dataset fills a gap in labeled 3D datasets and generation for training and evaluating models and techniques for TDA.
\end{abstract}

\section{Introduction}
\begin{figure}[htbp]
    \centering
    \includegraphics[height=45mm]{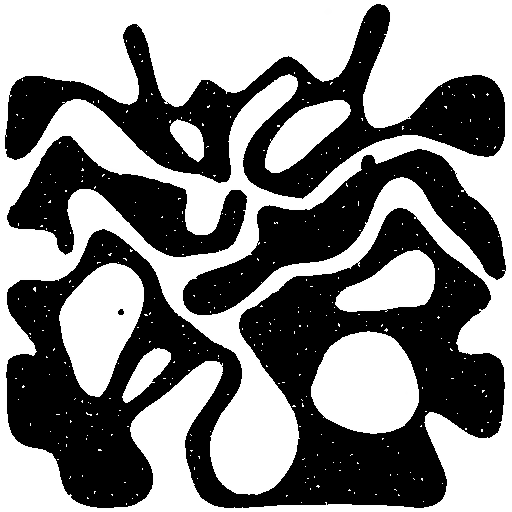}\\[1ex]
    \includegraphics[height=45mm]{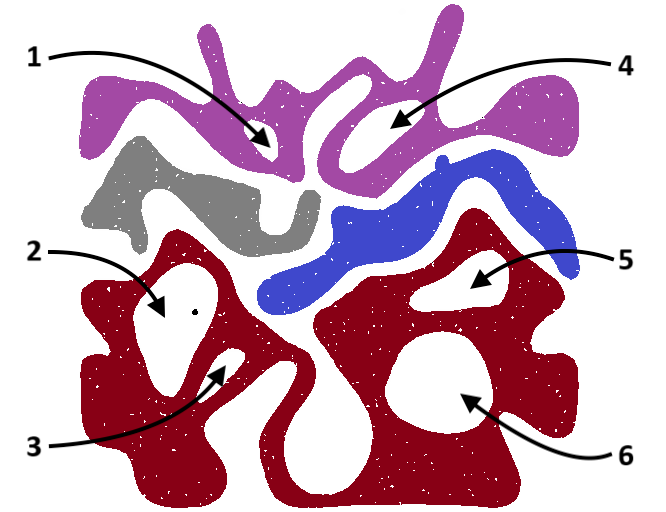}
    \caption{A 2D slice of a sample generated using the technique outlined in \cref{secgenerate}. The 2D binary image shows 6 holes across 4 disconnected objects. Top: raw sample. Bottom: annotated analysis.}
    \label{2dslice}
\end{figure}

\begin{figure*}[htbp]
\centerline{
{\includegraphics[trim=200mm 0 200mm 0, clip, height=45mm]{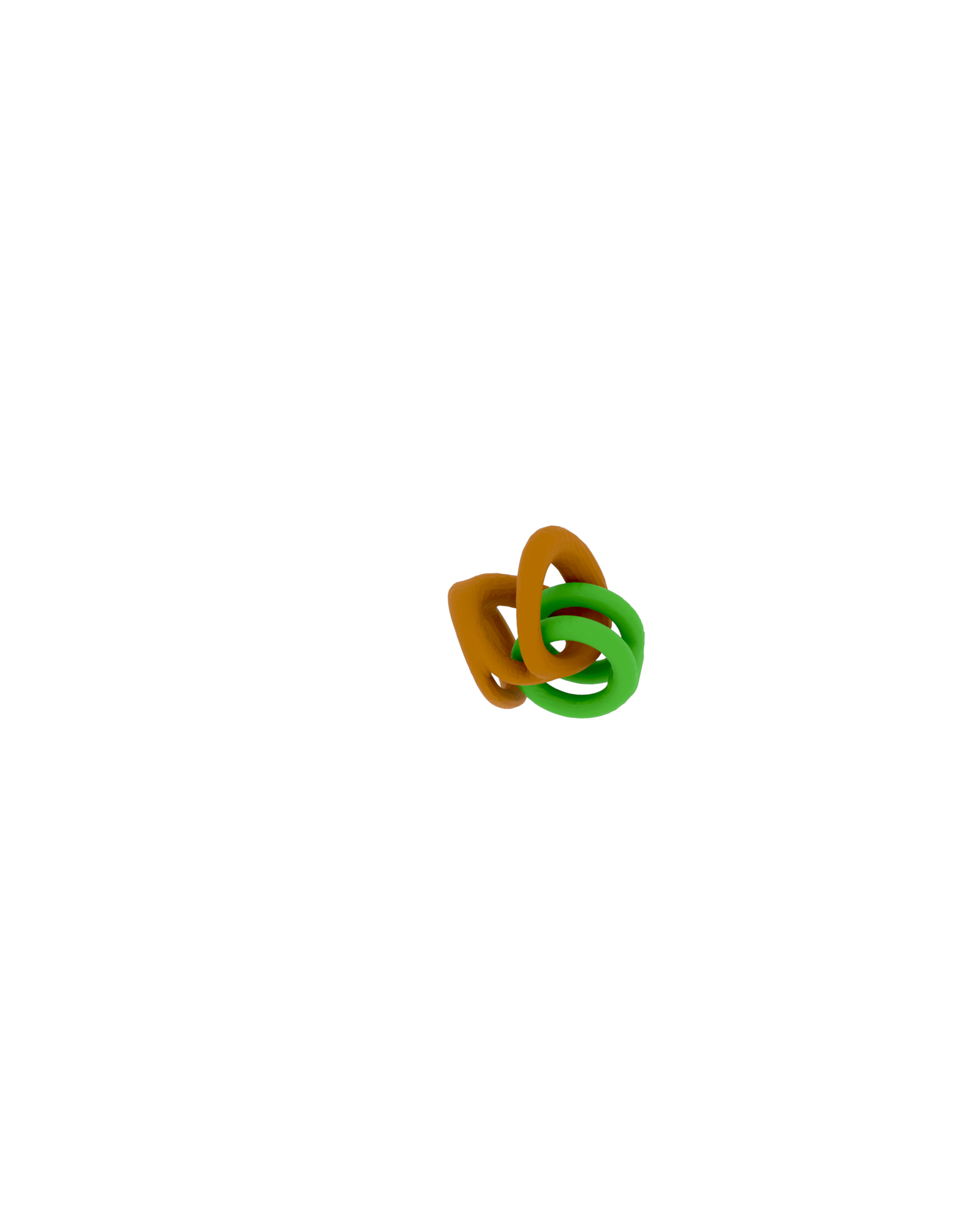}}
{\includegraphics[trim=0 0 100mm 0, clip,height=45mm]{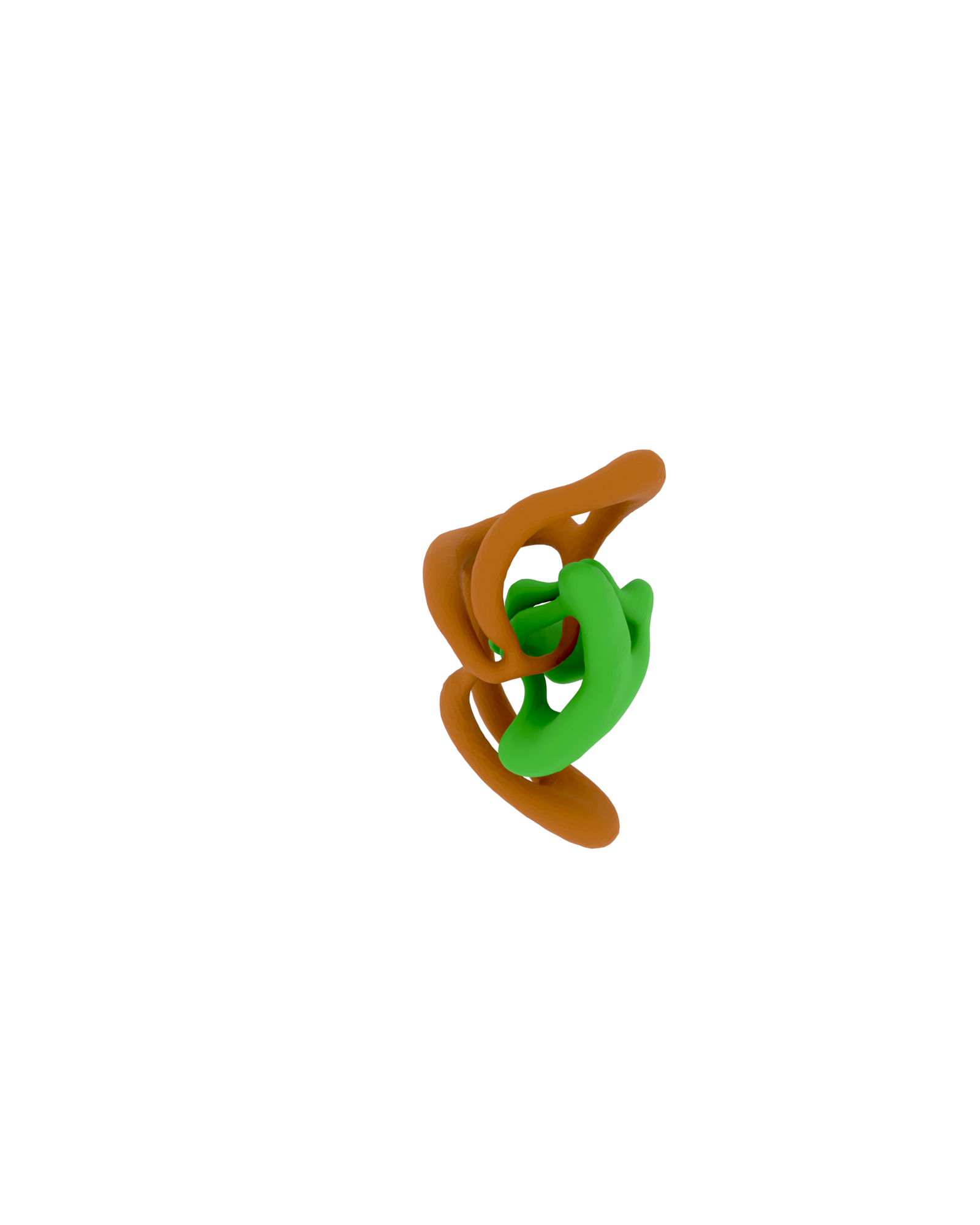}}
{\includegraphics[height=45mm,width=32mm]{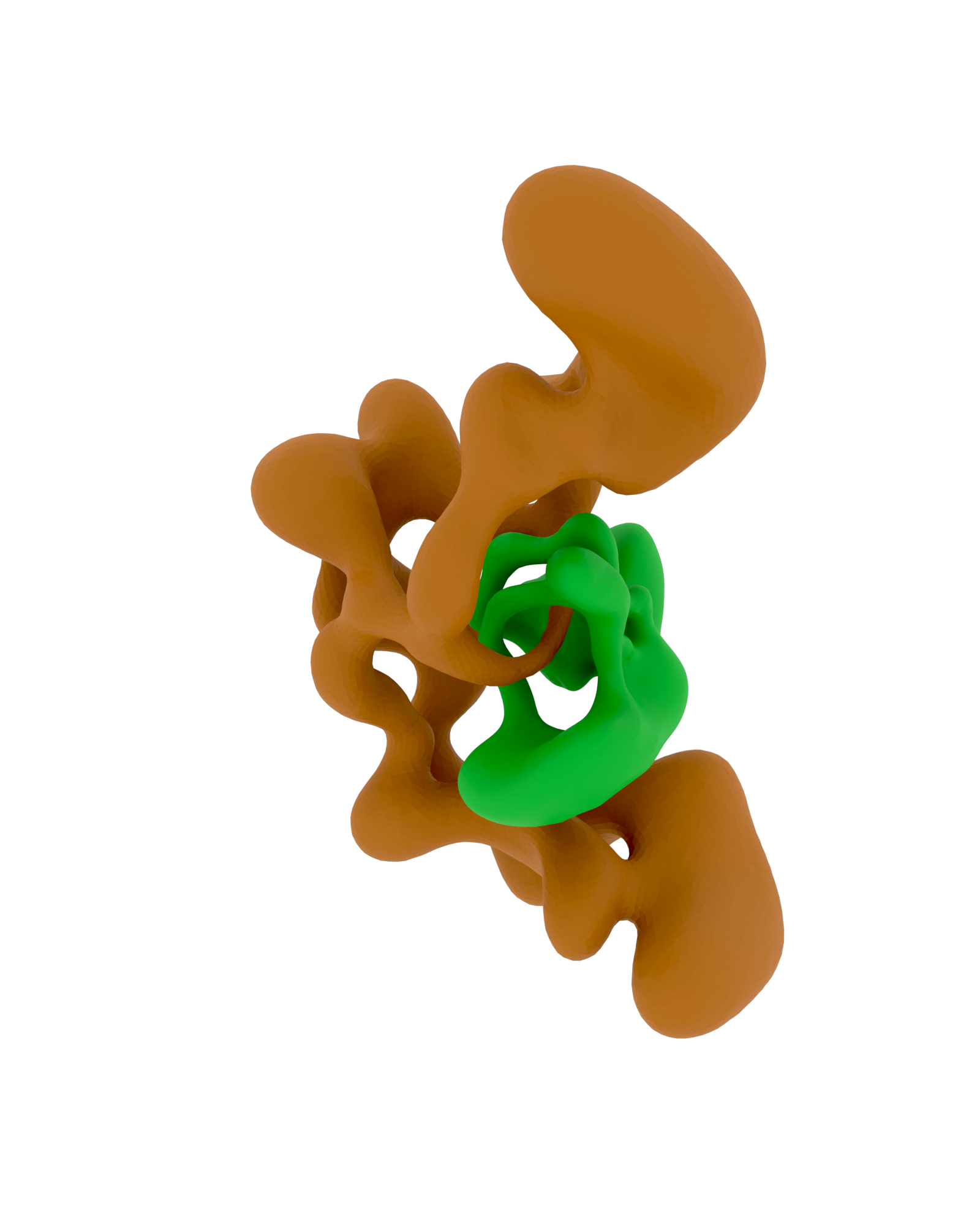}}
{\includegraphics[height=45mm,width=32mm]{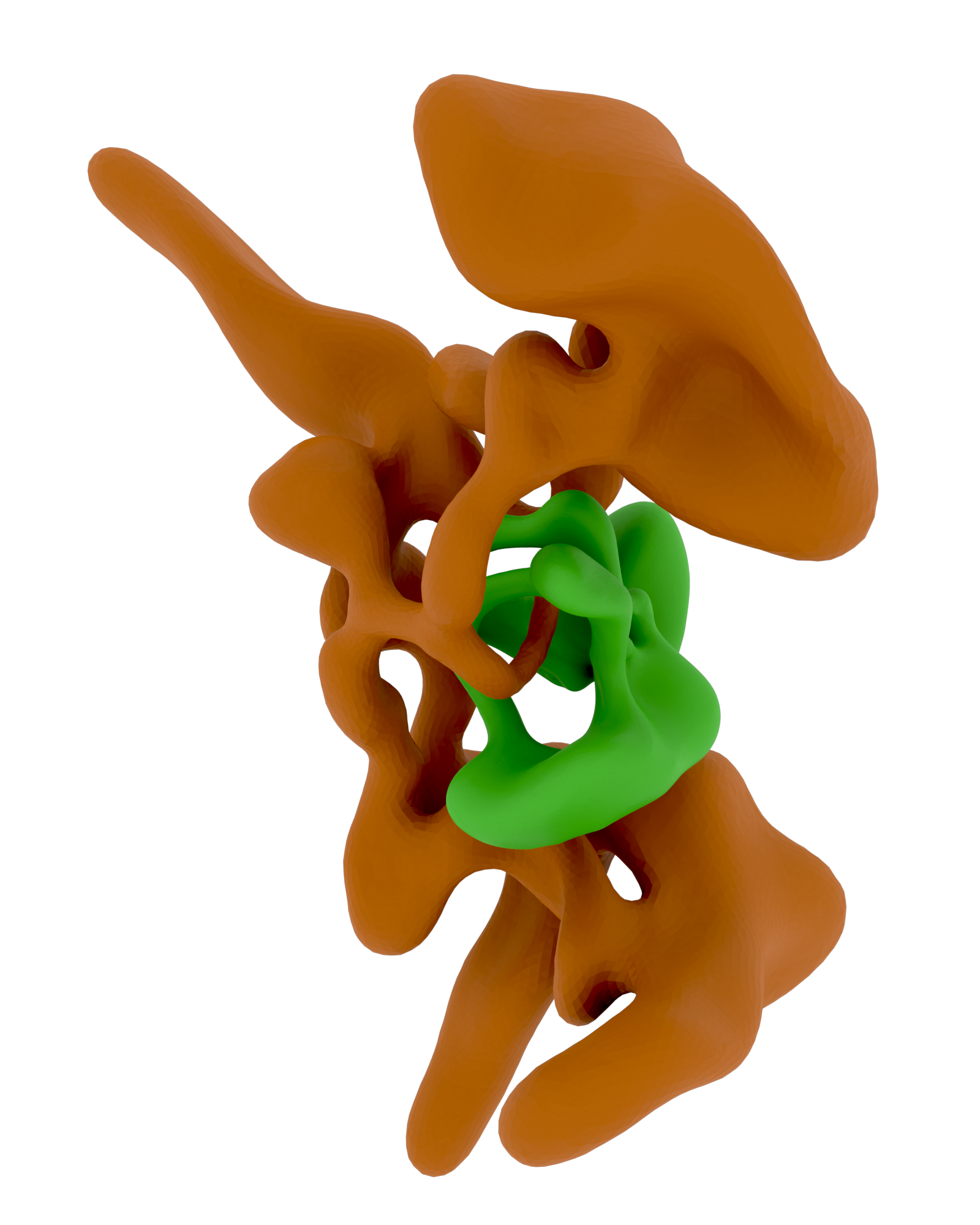}}}
\caption{Random growth of interlinked genus 2 (green) and genus 3 (brown) objects using the method outlined in \cref{secgenerate}. Visualisation performed in Blender 3.0.1.~\cite{blender3.0.1}}
\label{fig:datasetgrowth}
\end{figure*}

The shape of an object is an essential feature that can be used for classification. Topology is the branch of mathematics that formalizes the study of shapes through a rigorous analysis of connectivity, cavities, and holes within an object. Topological Data Analysis (TDA), a relatively new field at the intersection of mathematics, computer science, and data science, focuses on analyzing the shape of data or objects~\citep{Hatcher2002,EdelsbrunnerHarer2010}. TDA offers powerful tools, such as Persistent Homology (PH), for understanding complex data by extracting topological features like 
$n$-dimensional holes, which uncover underlying patterns and relationships.

While Persistent Homology is a powerful tool, it has certain drawbacks, such as high computational cost, particularly for large datasets. To address this, machine learning techniques have been proposed and successfully applied to extract topological features and signatures~\citep{paul2019estimating,HannouchChalup2023, PeekEtAl2023,de2022ripsnet}. The underlying idea behind these studies is to bypass the persistent homology computation and directly estimate or predict topological invariants, such as \textit{Betti numbers} and \textit{genus}, through neural networks to reduce computational cost and gain additional insights. Additionally, neural networks have been employed in various other TDA tasks, including analyzing outputs from traditional techniques like persistent homology for applications such as medical diagnosis and classification~\citep{yamanashi2021topological,rucco2014using,chung2021persistent}.

Approaches that bypass PH computation and directly estimate topological invariants typically require large datasets for training and testing. The availability of such extensive data repositories is a major bottleneck in the further development of this area. Our current work addresses this issue by proposing a new method to synthetically generate complex and versatile labeled datasets suitable for the training and testing of neural networks in topological classification.

An example of a visual approach to estimating hole counts in 2D data can be seen in \cref{2dslice}. This example is a 2D cross-section of the 3D generation process outlined in \cref{secgenerate}.

Our method utilizes the Repulsive Surfaces algorithm~\cite{yu2021repulsive} to perform homeomorphic (i.e., topology-preserving) deformations with randomized parameters and environmental constraints. This process generates a sequence of 3D data with known labels through an iterative growth mechanism, as illustrated in \cref{rgdatasetclouds}. This data generation approach allows for:
\begin{itemize}[itemsep=2pt, topsep=2pt] 

\item Incremental complexity in the generated data, making it suitable for both 3D or time-series 3D tasks. This also allows the accuracy of a model to be assessed at various geometrically complex stages while retaining the same topological complexities. 

\item Customizable growth configurations, allowing for adaptation to various applications and styles. 

\end{itemize}

Our method fills a critical gap in accessible, topologically labeled 3D/4D data and has been utilized to train neural networks for TDA tasks using `Betti Number' topological signatures (in a form we refer to as `genus' in this paper, see \cref{Background}). We demonstrate the efficacy of this synthetic data generation method through experiments with a 3D Convolutional Transformer Network (3DCTN)~\citep{3dctn}. The dataset used in these experiments is called the Random Grid Repulse Dataset, or simply `RG Repulse.' Details of the generation process are provided in \cref{rgexperiments}.

\subsection{Related Works}
Previous studies have used a variety of datasets and output structures to perform TDA on raw input data. One line of work has applied convolutional neural network architectures to estimate the Betti numbers of 2D and 3D data~\cite{paul2019estimating}. The 2D model was trained on image data consisting of randomly placed circles with randomized radii, and for the 3D model, this approach was scaled into 3D with random spheres embedded in a volumetric space. The concept has also been extended to 4D, where a synthetic dataset introduced higher-dimensional holes through cutouts from a solid 4D structure, with objects randomly scaled and rotated~\cite{HannouchChalup2023}.

Deep learning models have also been developed to estimate topological features directly from images \cite{som2020pi}. For example, Pi-Net was trained using datasets such as SVHN~\citep{netzer2011reading}, CIFAR10, and CIFAR100~\citep{krizhevsky2009learning}. The SVHN dataset features photos of house numbers, while CIFAR100 includes 100 common classes such as ship' or dog'.

Other efforts have focused on 3D point cloud TDA, with one approach sampling point clouds from the surface of objects in the ModelNet10 dataset~\citep{wu20153d} to train a RipsNet model designed for topological feature extraction \cite{de2022ripsnet}. A similar strategy was employed to train TopologyNet using the expanded ModelNet40 dataset, which includes 40 object classes instead of 10 \cite{zhou2022learning}.

A key limitation across these studies is dataset availability. Datasets like CIFAR100, SVHN, and ModelNet40 contain strong correlations between the geometrical and topological properties of the objects. For example, recognizing that an object is a `mug' inherently provides information about its toroidal topology. This correlation makes it difficult for networks to learn topological features independently of geometrical ones, potentially leading to overfitting. Additionally, these datasets possess fewer hole counts and limited topological complexity, restricting the variety of topological features. Finally, the distribution of samples in these datasets is optimized for object classification, not for balancing topological features like Betti numbers.

 A previous study introduced the WFC Repulse dataset \cite{PeekEtAl2023}, which aimed to mitigate these limitations by generating synthetic data with varying topological complexity. This approach involved creating 3D scenes with multiple objects, each having a different genus (number of $\beta_1$ holes). This method allowed for controlled topological complexity, which was used to train transformer networks for TDA.

This paper builds upon the WFC Repulse dataset by expanding and refining the generation process. The differences include the selection of generation parameters, environment generation, post-processing, data format and hole counts. The previous method used uniform mesh-surface sampling in 4096 point clouds, while the proposed method converts the meshes to voxel cubes to better simulate use cases such as medical scans or material science scans. This voxel data can then be treated as a volumetric cube (3D voxel data) or converted to a point cloud for common neural network architecture pipelines. 

\section{Background in Topology}\label{Background}

In this section we will briefly recall the basic notions in topology that are used in TDA and then describe the standard pipeline of Persistent Homology computation. For further reading see \citep{Hatcher2002, EdelsbrunnerHarer2010}.

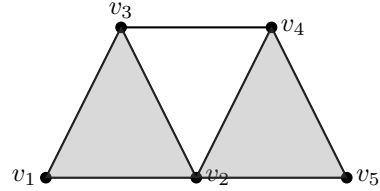
\begin{figure}
    \begin{center}
\begin{tikzpicture}

\filldraw[black] (0, 0) circle (2pt) node[anchor=east] {$v_1$};
\filldraw[black] (2, 0) circle (2pt) node[anchor=west] {$v_2$};
\filldraw[black] (1, 2) circle (2pt) node[anchor=south] {$v_3$};
\filldraw[black] (3, 2) circle (2pt) node[anchor=west] {$v_4$};
\filldraw[black] (4, 0) circle (2pt) node[anchor=west] {$v_5$};

\draw[thick] (0,0) -- (2,0) node[midway, below] {};
\draw[thick] (0,0) -- (1,2) node[midway, left] {};
\draw[thick] (2,0) -- (1,2) node[midway, right] {};
\draw[thick] (2,0) -- (4,0) node[midway, below] {};
\draw[thick] (2,0) -- (3,2) node[midway, right] {};
\draw[thick] (3,2) -- (4,0) node[midway, right] {};
\draw[thick] (1,2) -- (3,2) node[midway, above] {};

\filldraw[gray, opacity=0.3] (0,0) -- (2,0) -- (1,2) -- cycle;
\filldraw[gray, opacity=0.3] (2,0) -- (4,0) -- (3,2) -- cycle;

\end{tikzpicture}
\end{center}
    \caption{Example of a simplicial complex}
    \label{fig:simpcompexp}
\end{figure}

A geometric $k$-simplex is the convex hull of $k+1$ affinely independent points in $\mathbf{R}^d$. For example, a point is a $0$-simplex, an edge is a $1$-simplex, a triangle is a $2$-simplex, and a tetrahedron is a $3$-simplex. A subset simplex is called a face of the original simplex. A geometric simplicial complex is a collection of geometric simplices that intersect only at their common faces and are closed under the face relation. See \cref{fig:simpcompexp}
for an example. A filtration or a filtered simplicial complex is a sequence
of nested simplicial complexes indexed by a scale parameter.\\
Homology, is a tool from algebraic topology to quantify the number of $k$-dimensional topological features (or holes) in a topological space, such as a simplicial complex. For instance, $H_0$ the zero degree homology classes describe the number of connected components, $H_1$ the one degree homology classes describe the number of loops, and $H_2$, the two dimensional homology classes quantifies voids or cavities. The ranks of these homology groups are referred to as Betti numbers. In particular, the Betti number $\beta_ k$ corresponds to the number of $k$-dimensional holes. In \cref{fig:simpcompexp} the Betti numbers are as follows, $\beta_0 =1, \beta_1 = 1, \text{ and } \forall k \geq 2, \beta_k = 0$. \cref{table:bettis} shows the list of non-zero Betti numbers for all connected compact oriented 2-manifolds~\citep{SeifertThrelfall1934} along with other topological invariants such as \textit{genus} $g$ and \textit{Euler Characteristic} \(\chi = \beta_0 - \beta_1 + \beta_2 \), which in this case are related as follows \(g = \frac{\beta_1}{2} = -\frac{\chi-2}{2}\).

\begin{table}[ht]
\vspace{2ex}
\centering
\caption{Genus ($g$), Betti numbers ($\beta_n$), and Euler characteristic ($\chi$) of closed compact orientable surfaces.} 
\label{background:topology:table}
\begin{tabular}{lccccc}
\toprule 
Surface $M$ &$g$&$\beta_0$&$\beta_1$&$\beta_2$&$\chi$\\[3pt]\midrule
Sphere  $S^2$&0&1&0&1&2\\
Torus $T^2$&1&1&2&1&0\\
$g$-holed torus $T^2\# \dots \# T^2$&$g$&1&2$g$&1&2-2$g$\\[3pt]\bottomrule
\end{tabular}\label{table:bettis}
\end{table}

In a standard pipeline of \textit{persistent homology} computation, a dataset, typically represented as a point cloud in a metric space, is used to first build a filtered simplicial complex which is used to construct a \textit{boundary matrix} which is then finally reduced in a special form to read off the persistent homology. A common method for constructing such filtered simplicial complex is the \textit{Vietoris-Rips} complex. This complex is formed by connecting data points that lie within a certain distance from each other, progressively increasing the complexity of the structure as the distance threshold grows.
The idea of filtration is used to analyze data across multiple scales. As the filtration parameter (here the distance threshold) increases, new simplices are added, enabling the tracking of how topological features, such as Betti numbers, emerge and disappear across different scales. \\

\section{Synthetic Data Generation} \label{secgenerate}
\begin{figure*}[htbp]
    \centering
    \hfill
    \subfloat[\label{fig:rg_seed5}]{
        \includegraphics[height=40mm]{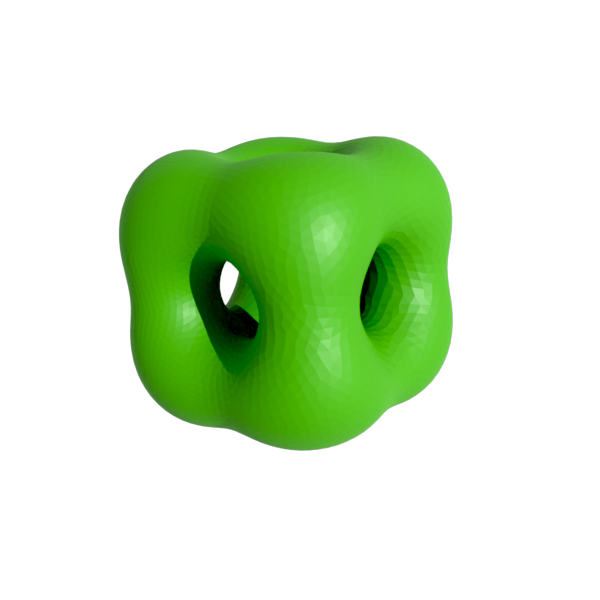}
    }
    \hfill
    \subfloat[\label{fig:rg_environment}]{
        \includegraphics[height=40mm]{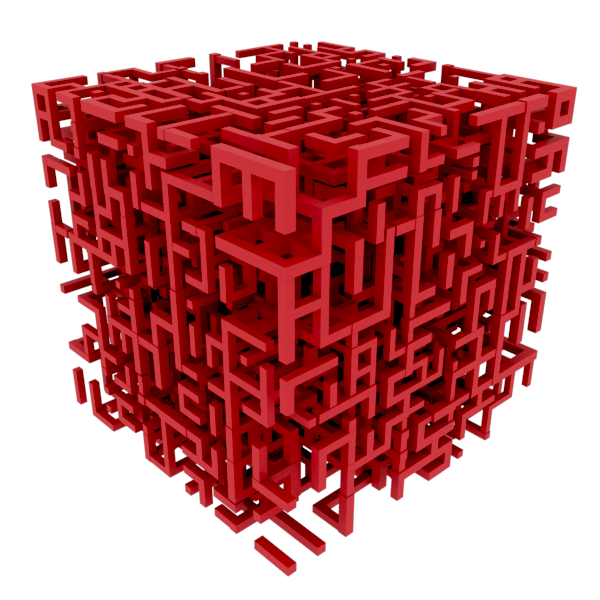}
    }
    \hfill
    \subfloat[\label{fig:genus5}]{
        \includegraphics[height=40mm,trim=0 50 0 0]{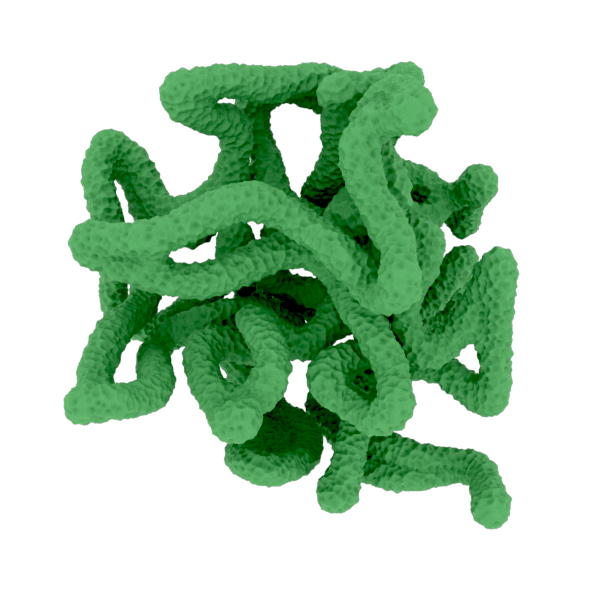}
    }
    \hfill{}
    \caption{Process from structure to synthetic object: (a) shows a genus 5 seed structure; (b) demonstrates an environment generated using the random grid method; and (c) displays the final genus 5 object, represented as a mesh with Voronoi surface displacement mapping.}
    \label{fig:full_generation_pipeline}
\end{figure*}

In a previous study,~\cite{PeekEtAl2023} created a synthetic dataset called the WFC Repulse dataset using the Wave Function Collapse algorithm \citep{gumin2016wavefunctioncollapse} and Repulsive Surfaces algorithm \citep{yu2021repulsive}. This study demonstrated the viability of the proposed data generation method for topological training and assessed the segmentation ability of neural networks using topological signatures. Building on that, we describe how this method has been altered to produce a new RG Dataset.

We begin by outlining the basic steps of the data generation process. This process can be adapted to generate various types of datasets depending on the learning task.

\begin{enumerate}[itemsep=1pt, topsep=1pt]
    \item \textit{Seed}: This step manually creates a `seed' object in the form of a 3D mesh with known topology.
    \item \textit{Environment}: Next, a randomly generated `environment' was generated to grow the seed within. This environment will act as a constraint on which the seed will grow.
    \item \textit{Scene setup}: Then, the seed is randomly placed inside the environment using random placement and scaling.
    \item \textit{Deformations}: Next, the Repulsive Surfaces algorithm is used increase the seeds surface area within the environmental constraint. This process deforms the seed object without altering its topology.
    \item  \textit{Subsampling}: Finally, the grown seed can be converted into a voxel or point cloud form with various noise/scaling applied. 
\end{enumerate}

The approach outlined above was common in both the previous study's WFC dataset~\citep{PeekEtAl2023} and the new RG Dataset. The key differences include the hole count which was raised from $[0-3]$ to $[0-20]$. The WFC Dataset was processed to a point cloud format via mesh surface sampling, while the RG Dataset was aimed to emulate 3D pixel data such as medical or material science scans. This involved voxelization and noise additions. The environment generation method was also changed from the wave function collapse algorithm to a new random grid approach. The random grid process is outlined below and the Wave Function Collapse algorithm can be seen in \cref{appendix:wfc}. The creation of a random grid allowed more flexibility and control in the scale, thickness and density of sections and used real number parameters for placement and thickness over discrete tile cells.

Now, we will explain each of the above outlined steps in more detail to generate the current RG Repulse dataset.

\paragraph{Seed} We manually created 21 seed meshes for the RG Repulse dataset. Each of these meshes has a different number of 1-dimensional holes ($\beta_1$, genus), ranging from $0$ to $20$. See \cref{fig:rg_seed5} for an example genus 5 seed.

\begin{figure*}[htbp]
    \setlength{\fboxrule}{1pt} %
    \setlength{\fboxsep}{0pt}  %
    \centering
    \fbox{\includegraphics[width=0.18\textwidth]{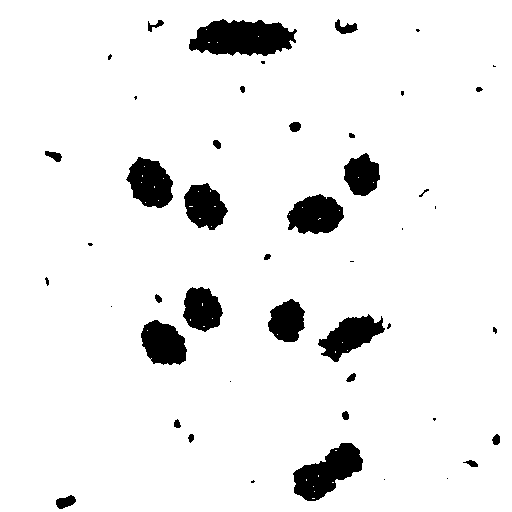}}
    \fbox{\includegraphics[width=0.18\textwidth]{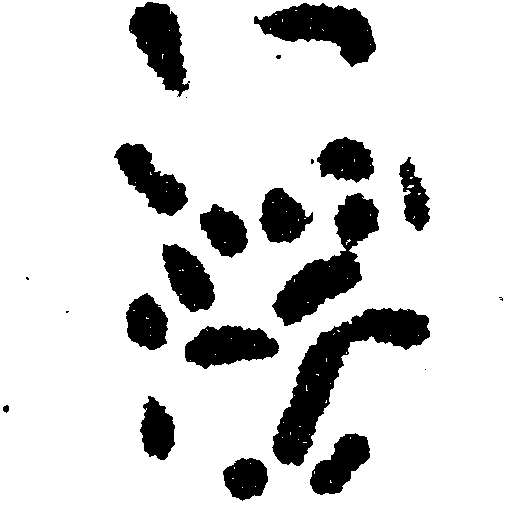}}
    \fbox{\includegraphics[width=0.18\textwidth]{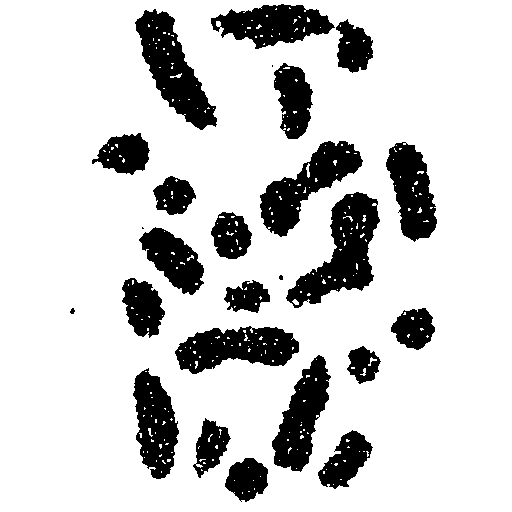}}
    \fbox{\includegraphics[width=0.18\textwidth]{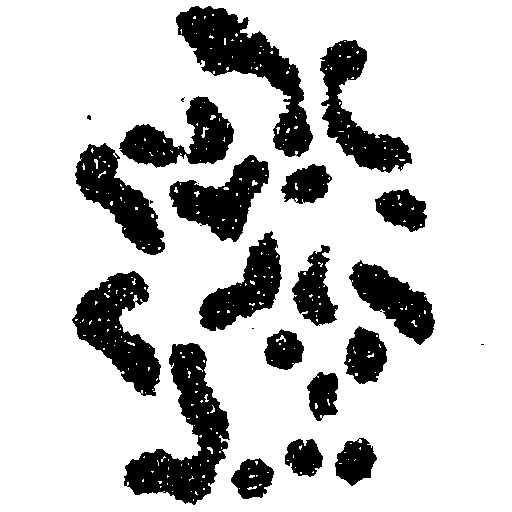}}
    \fbox{\includegraphics[width=0.18\textwidth]{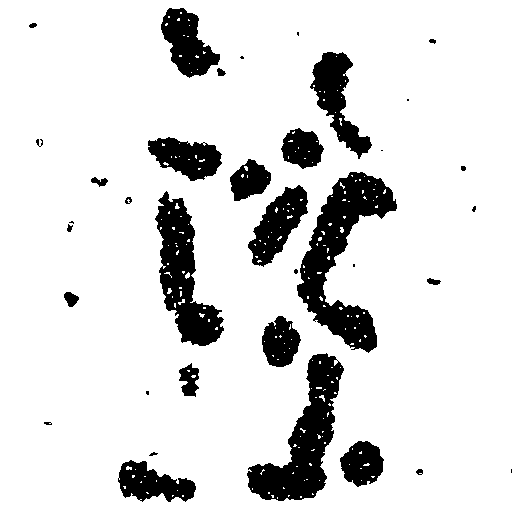}}

    \caption{Cross-sectional slices of a genus 5 object with Voronoi mesh displacement mapping and 3 octaves of 3D Perlin noise.}
    \label{fig:RGslices}
\end{figure*}

\paragraph{Random Grid Environment } A unique environment is generated for every grown seed in the RG dataset. As the seed is grown within the environment, the geometry of the environment will determine aspects of the grown samples. We start this generation by dissecting a cube with a side length of 20 into \(5^3\) smaller cubic chunks, each with a side length of 4. Each of these smaller chunks is assigned different randomized parameters to create distinct sub-regions in the environment. Having different regions will produce geometric diversity as the sample in constrained in different ways. 
These random parameters include:

\textit{Axis resolution}: A random resolution between 2 and 4 is selected for each axis in the subchunk, denoted as $\mathbf{x_{res}}$, $\mathbf{y_{res}}$, and $\mathbf{z_{res}}$. This produces a 3D grid of points comprising $\mathbf{x_{res}} \times \mathbf{y_{res}} \times \mathbf{z_{res}}$ points for each subchuck separately.

\textit{Connection probability}: For each subchunk we assign a probability \(P\), randomly chosen between 0.15 and 0.25, referred to as the \textit{edge connection probability}. Adjacent points are connected by edges with the probability \(P\). This creates varied density in the overall environment as different subchuncks have different \(P\)s.

\textit{Edge thickness}: For each subchunk we assign a thickness \(T\), chosen randomly between 0.4 and 0.6. Finally, we mesh the 3D environment by adding rectangular prisms around each edge, with thickness corresponding to the subchunks assigned value of \(T\). 

An example of the generated environment is shown in \cref{fig:rg_environment}.

\paragraph{Scene Setup } Once the seed 3D mesh and the random 3D grid environment are ready, we place the seed meshes inside the environment. This is done by offsetting the bounding-box center of the seed to the environment's center. Then, the seed is \textit{rotated} randomly between $0$ and $2\pi$ around each axis. We perform an initial \textit{global scaling}  to adjust the surface area of the seed mesh to $1$, followed by an additional random scaling of $\pm 25\%$. We then apply another set of anisotropic scaling of $\pm 50\%$ to each axis independently.

These augmented steps introduce variability to the initial seed conditions, allowing for further diversity in the grown samples.    

\paragraph{Deformations } In this step, we aim to induce geometric complexity by forcing the surface area to increase within randomized constrained environments. To achieve this, we use the \textit{Repulsive Surfaces (RS) algorithm}~\citep{yu2021repulsive}. Intuitively, this is similar to growing a simpler `seed' mesh into a larger, more complex form through iterative \textit{homeomorphic} deformations. Typically, the RS algorithm is used to deform a complex manifold into a simpler equivalent form; however, in our approach, we reverse this goal and use it to add more complexity. The algorithm works by pushing apart pairs of points in an attempt to maintain a uniform distribution. This is combined with energy minimization, which is determined by the relationship between a points spacial distance and surface distance (curved surfaces have shorter pathways in spatial coordinates than surface coordinates), which penalizes objects with greater variance, see \citep{yu2021repulsive}. 

Since the generated sample is in mesh form, many surface mapping deformations can be applied. We used Blender~\citep{blender3.0.1} to apply a Voronoi displacement map, a method that subdivides a surface into regions based on proximity to a set of points, resulting in a distinct, organic-looking pattern. Voronoi is particularly useful for creating unique surface geometries, adding variation and breaking up the smoothness of the mesh. This displacement was applied with an intensity of 0.5 and a size of 0.1 to introduce surface variations, centered around the midlevel of the mesh. This approach was chosen to ensure each mesh has a unique, textured appearance, though other texture maps or real-world images could be used depending on the desired application.
 
\begin{figure}[htbp]    
\centering    
    \includegraphics[width=0.31\linewidth, trim=90 90 90 60, clip]{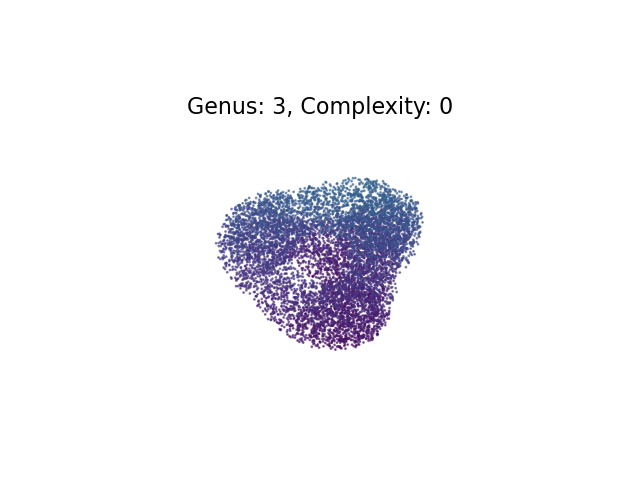}
    \includegraphics[width=0.31\linewidth, trim=90 90 90 60, clip]{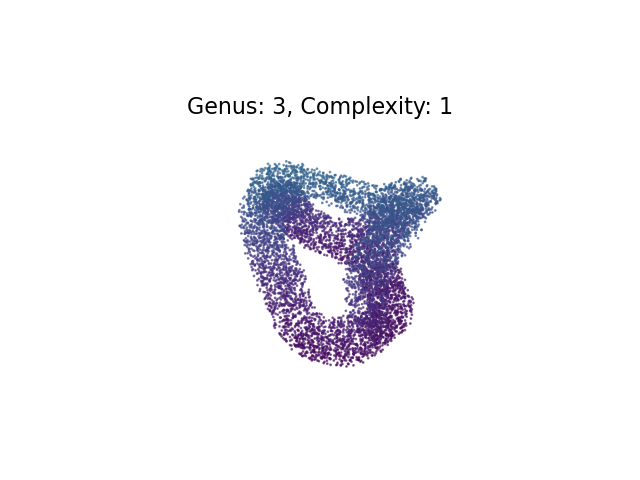}
    \includegraphics[width=0.31\linewidth, trim=90 90 90 60, clip]{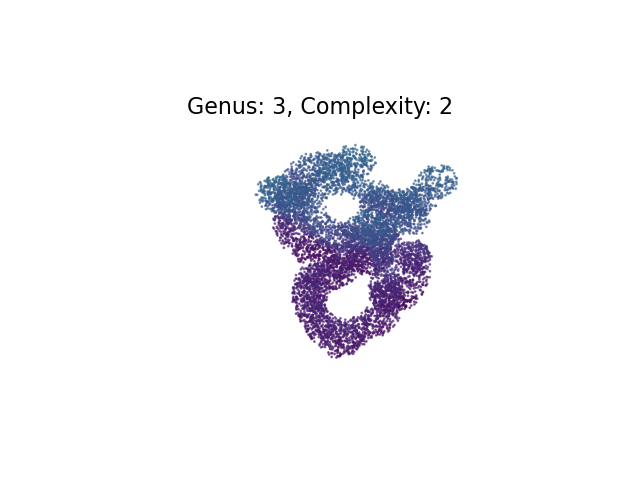}
    
    \vspace{0.5cm}
    
    \includegraphics[width=0.31\linewidth, trim=90 90 90 60, clip]{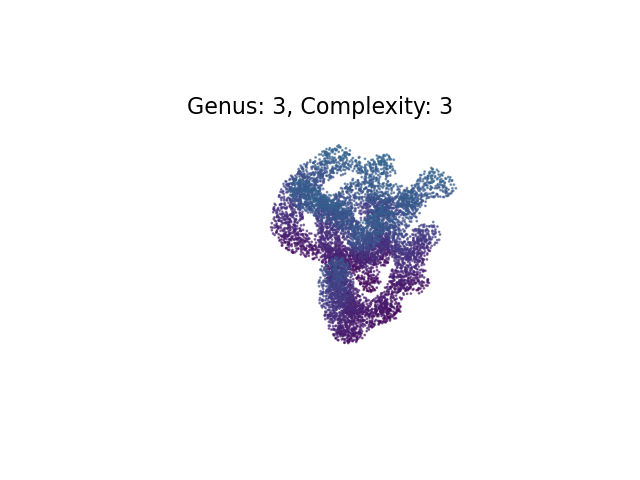}
    \includegraphics[width=0.31\linewidth, trim=90 90 90 60, clip]{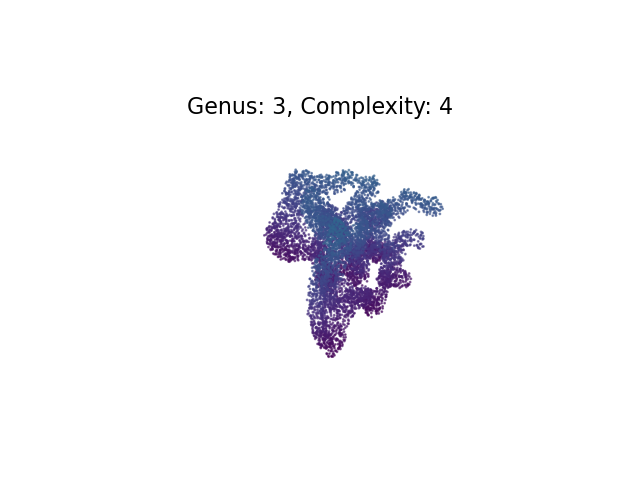}
    \includegraphics[width=0.31\linewidth, trim=90 90 90 60, clip]{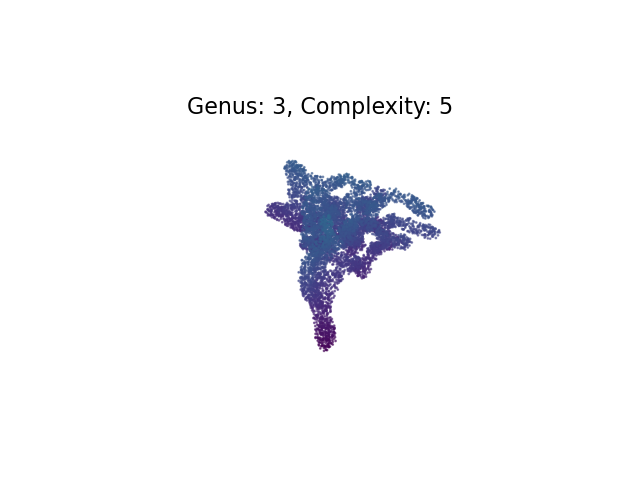}
    
    \caption{Examples of complexity levels for a genus 3 sample in the RG dataset. Each sample contains 8192 points. Complexity levels $[0-5]$ for objects of genus 10 can also be seen in \cref{appendix:complexity}.}
    \label{rgdatasetclouds}
\end{figure}

\begin{table*}[ht]
\centering
\caption{Class (hole count) vs complexity accuracy (\%) of 3DCTN hole count estimation.}
\label{rg:results}
\begin{tabular}{c|cccccc}
\toprule
Class & 0 & 1 & 2 & 3 & 4 & 5 \\
\midrule
0 & \cellcolor[HTML]{99FF99}100.00\% & \cellcolor[HTML]{A6FF99}93.33\% & \cellcolor[HTML]{B4FF99}86.67\% & \cellcolor[HTML]{DDFF99}66.67\% & \cellcolor[HTML]{DDFF99}66.67\% & \cellcolor[HTML]{EAFF99}60.00\% \\
1 & \cellcolor[HTML]{99FF99}100.00\% & \cellcolor[HTML]{C1FF99}80.00\% & \cellcolor[HTML]{EAFF99}60.00\% & \cellcolor[HTML]{FFEA99}40.00\% & \cellcolor[HTML]{FFEA99}40.00\% & \cellcolor[HTML]{FFC199}20.00\% \\
2 & \cellcolor[HTML]{B1FF99}87.50\% & \cellcolor[HTML]{D8FF99}68.75\% & \cellcolor[HTML]{E5FF99}62.50\% & \cellcolor[HTML]{FEFF99}50.00\% & \cellcolor[HTML]{F1FF99}56.25\% & \cellcolor[HTML]{FFF299}43.75\% \\
3 & \cellcolor[HTML]{99FF99}100.00\% & \cellcolor[HTML]{EAFF99}60.00\% & \cellcolor[HTML]{FFD599}30.00\% & \cellcolor[HTML]{FFAD99}10.00\% & \cellcolor[HTML]{FFAD99}10.00\% & \cellcolor[HTML]{FFAD99}10.00\% \\
4 & \cellcolor[HTML]{99FF99}100.00\% & \cellcolor[HTML]{A8FF99}92.31\% & \cellcolor[HTML]{D7FF99}69.23\% & \cellcolor[HTML]{FFF799}46.15\% & \cellcolor[HTML]{FFD799}30.77\% & \cellcolor[HTML]{FFE799}38.46\% \\
5 & \cellcolor[HTML]{99FF99}100.00\% & \cellcolor[HTML]{A4FF99}94.44\% & \cellcolor[HTML]{F3FF99}55.56\% & \cellcolor[HTML]{FFF399}44.44\% & \cellcolor[HTML]{FFE899}38.89\% & \cellcolor[HTML]{FFF399}44.44\% \\
6 & \cellcolor[HTML]{99FF99}100.00\% & \cellcolor[HTML]{BDFF99}81.82\% & \cellcolor[HTML]{E3FF99}63.64\% & \cellcolor[HTML]{F5FF99}54.55\% & \cellcolor[HTML]{F5FF99}54.55\% & \cellcolor[HTML]{F5FF99}54.55\% \\
7 & \cellcolor[HTML]{A8FF99}92.31\% & \cellcolor[HTML]{B8FF99}84.62\% & \cellcolor[HTML]{F7FF99}53.85\% & \cellcolor[HTML]{FFD799}30.77\% & \cellcolor[HTML]{FFC899}23.08\% & \cellcolor[HTML]{FFC899}23.08\% \\
8 & \cellcolor[HTML]{99FF99}100.00\% & \cellcolor[HTML]{99FF99}100.00\% & \cellcolor[HTML]{CBFF99}75.00\% & \cellcolor[HTML]{CBFF99}75.00\% & \cellcolor[HTML]{FEFF99}50.00\% & \cellcolor[HTML]{CBFF99}75.00\% \\
9 & \cellcolor[HTML]{DDFF99}66.67\% & \cellcolor[HTML]{DDFF99}66.67\% & \cellcolor[HTML]{FFDD99}33.33\% & \cellcolor[HTML]{FFDD99}33.33\% & \cellcolor[HTML]{FFDD99}33.33\% & \cellcolor[HTML]{FFBA99}16.67\% \\
10 & \cellcolor[HTML]{C8FF99}76.92\% & \cellcolor[HTML]{B8FF99}84.62\% & \cellcolor[HTML]{B8FF99}84.62\% & \cellcolor[HTML]{D7FF99}69.23\% & \cellcolor[HTML]{F7FF99}53.85\% & \cellcolor[HTML]{FFF799}46.15\% \\
11 & \cellcolor[HTML]{99FF99}100.00\% & \cellcolor[HTML]{99FF99}100.00\% & \cellcolor[HTML]{F7FF99}53.85\% & \cellcolor[HTML]{FFE799}38.46\% & \cellcolor[HTML]{FFC899}23.08\% & \cellcolor[HTML]{FFB899}15.38\% \\
12 & \cellcolor[HTML]{99FF99}100.00\% & \cellcolor[HTML]{99FF99}100.00\% & \cellcolor[HTML]{B8FF99}84.62\% & \cellcolor[HTML]{D7FF99}69.23\% & \cellcolor[HTML]{E7FF99}61.54\% & \cellcolor[HTML]{FFF799}46.15\% \\
13 & \cellcolor[HTML]{99FF99}100.00\% & \cellcolor[HTML]{99FF99}100.00\% & \cellcolor[HTML]{99FF99}100.00\% & \cellcolor[HTML]{CBFF99}75.00\% & \cellcolor[HTML]{CBFF99}75.00\% & \cellcolor[HTML]{EDFF99}58.33\% \\
14 & \cellcolor[HTML]{99FF99}100.00\% & \cellcolor[HTML]{99FF99}100.00\% & \cellcolor[HTML]{99FF99}100.00\% & \cellcolor[HTML]{FFE399}36.36\% & \cellcolor[HTML]{FFF599}45.45\% & \cellcolor[HTML]{FFE399}36.36\% \\
15 & \cellcolor[HTML]{99FF99}100.00\% & \cellcolor[HTML]{99FF99}100.00\% & \cellcolor[HTML]{99FF99}100.00\% & \cellcolor[HTML]{99FF99}100.00\% & \cellcolor[HTML]{ABFF99}90.91\% & \cellcolor[HTML]{D0FF99}72.73\% \\
16 & \cellcolor[HTML]{99FF99}100.00\% & \cellcolor[HTML]{99FF99}100.00\% & \cellcolor[HTML]{C1FF99}80.00\% & \cellcolor[HTML]{EAFF99}60.00\% & \cellcolor[HTML]{EAFF99}60.00\% & \cellcolor[HTML]{FEFF99}50.00\% \\
17 & \cellcolor[HTML]{99FF99}100.00\% & \cellcolor[HTML]{99FF99}100.00\% & \cellcolor[HTML]{99FF99}100.00\% & \cellcolor[HTML]{C4FF99}78.57\% & \cellcolor[HTML]{D3FF99}71.43\% & \cellcolor[HTML]{D3FF99}71.43\% \\
18 & \cellcolor[HTML]{ADFF99}90.00\% & \cellcolor[HTML]{99FF99}100.00\% & \cellcolor[HTML]{ADFF99}90.00\% & \cellcolor[HTML]{EAFF99}60.00\% & \cellcolor[HTML]{FEFF99}50.00\% & \cellcolor[HTML]{FFD599}30.00\% \\
19 & \cellcolor[HTML]{99FF99}100.00\% & \cellcolor[HTML]{99FF99}100.00\% & \cellcolor[HTML]{99FF99}100.00\% & \cellcolor[HTML]{ABFF99}90.91\% & \cellcolor[HTML]{ABFF99}90.91\% & \cellcolor[HTML]{BDFF99}81.82\% \\
20 & \cellcolor[HTML]{EDFF99}58.33\% & \cellcolor[HTML]{DDFF99}66.67\% & \cellcolor[HTML]{EDFF99}58.33\% & \cellcolor[HTML]{EDFF99}58.33\% & \cellcolor[HTML]{FFED99}41.67\% & \cellcolor[HTML]{FFED99}41.67\% \\
\bottomrule
\end{tabular}
\end{table*}

\paragraph{Subsampling } The final mesh, after all deformations, was sampled into a 3D voxel space with a 
\(256^3\) resolution. 
We then generated three octaves of Perlin noise, a gradient noise algorithm commonly used in procedural texture generation, and applied them to the voxel space. Perlin noise creates smooth, continuous variations, which are particularly effective for adding natural-looking randomness to the voxel grid. Each octave represents a layer of noise with different frequencies and intensities, allowing for more complex surface deformations.
In this context, the \textit{scale} adjusts the size of the noise patterns, with lower values producing larger, more prominent features. The \textit{threshold} defines which parts of the noise should be considered significant enough to alter the voxel space.
The first octave using a scale of 4 and threshold of 0.5 which was added to the cube. 
The second octave using a scale of 8 and threshold of 0.55 which was added to the cube. 
The third octave using a scale of 16 and threshold of 0.55 which was subtracted from the cube. 
Afterwards, a \textit{smoothing} step was performed by applying a Gaussian filter to the 3D voxel data with a standard deviation 
\(\sigma = 0.25\) to achieve a smoother representation.
These noise, scale, and resolution settings are easily configurable for different objectives. An example result is illustrated in \cref{fig:RGslices}. These 2D slices are cross-sections of the sample shown in \cref{fig:genus5}. The generation time of these samples can be seen in \cref{rg:generation_time}. Next, we provide the specifications for the final generated RG Repulse data.

\paragraph{RG Dataset Specifications}
The 21 seed meshes, each consisting of 1-dimensional holes ranging from \(0\) to \(20\), were grown in unique random environments with surface area increases ranging from 15--20x. Complexity levels between 0 and 5 were assigned at each 20\% of the samples overall growth. See \cref{rgdatasetclouds} for example complexity levels. These samples were then converted into voxel cubes with a \(256^3\) resolution. Finally, the RG dataset comprises 6,366 training samples and 1,456 test samples.

\section{Experimental Results}\label{rgexperiments}

We used our labeled dataset `RG Repulse,' generated using the algorithm described in the previous section, to train a 3D Convolutional Transformer Network (3DCTN, ~\cite{3dctn}) for topological classification based on Betti numbers, specifically \(\beta_1\). This section outlines the experimental results regarding the model training and performance. The parameters used for training are detailed in \cref{rg:parameters}.

\paragraph{Results }
The \(256^3\) resolution voxel cubes in the RG dataset were uniformly sampled into 8,192 points to generate a sparser point cloud. This subsampling requires fewer computational resources than the original 16 million points. The 3DCTN results on the test set are shown in \cref{rg:results}.

\begin{table}[ht] 
\centering
\caption{Training parameters}\label{rg:parameters}
\begin{tabular}{c c | c c}
\toprule
\textbf{Parameter} & \textbf{Value} & \textbf{Parameter} & \textbf{Value} \\
\midrule
Model & 3DCTN & Learning rate & 0.01 \\
Optimizer & SGD & Weight decay & 0.0001 \\
Point count & 8192 & Epoch & 300 \\
\bottomrule
\end{tabular}
\end{table}

\section{Discussion and Conclusion}
The results of the RG experiment demonstrated a decrease in accuracy as complexity levels increased. This suggests that greater homeomorphic deformation introduces more variability and challenge within the samples. Such variability is ideal for training neural networks for TDA tasks, which are invariant to geometric differences. The diversity in appearance across topologically equivalent samples can help mitigate overfitting. Additionally, this dataset has potential utility beyond machine learning models, including for the evaluation of persistent homology algorithms.

Previous experiments~\citep{PeekEtAl2023} conducted on the WFC dataset—which featured samples with lower hole counts, smoother surfaces, and point cloud formats—demonstrated the capability of transformers to segment data using topological labels. Per-point segmentation enables each object within a scene to be identified, classified, and localized. This is valuable because it allows for the preservation and analysis of relationships between metric, geometric, and topological properties such as size, volume, and shape.

By adjusting parameters like seeds, growth rates, sample sizes, hole sizes, and post-processing techniques, we gain substantial control over the appearance and properties of the generated samples. This flexibility is evident in the differences between the WFC and RG datasets. For real-world applications, this process can be tailored to replicate key aspects of actual data. Consequently, transfer learning can then be applied using smaller real-world datasets, provided that some understanding of topological structures exists.

TDA is a rapidly growing field with an increasing need for labeled datasets. The data generation method presented in this paper aims to address this gap by providing data that is both rich in topological variety and appropriately labeled. Our experiments demonstrate the viability of this dataset generation technique, as well as the inherent challenges associated with it. Conceptually, TDA presents different challenges compared to conventional classification tasks due to the significant visual differences between objects of the same topological class. This synthetic data allows researchers to explore and evaluate specific topological features without interference from extraneous variables that may be present in real-world data.

\subsection*{Acknowledgements} This research was supported by the Australian Government through the ARC's Discovery Projects funding scheme (project DP210103304). The first author was supported by a Research Training Program (RTP) Scholarship – Fee Offset by the Commonwealth Government.

{
    \small
    \bibliographystyle{ieeenat_fullname}
    \bibliography{literature.bib}

\begin{thebibliography}{19}
\providecommand{\natexlab}[1]{#1}
\providecommand{\url}[1]{\texttt{#1}}
\expandafter\ifx\csname urlstyle\endcsname\relax
  \providecommand{\doi}[1]{doi: #1}\else
  \providecommand{\doi}{doi: \begingroup \urlstyle{rm}\Url}\fi

\bibitem[Blender~Foundation(2021)]{blender3.0.1}
Blender Development~Team Blender~Foundation.
\newblock Blender 3.0.1, 2021.
\newblock Software.

\bibitem[Chung et~al.(2021)Chung, Hu, Lo, and Wu]{chung2021persistent}
Yu-Min Chung, Chuan-Shen Hu, Yu-Lun Lo, and Hau-Tieng Wu.
\newblock A persistent homology approach to heart rate variability analysis
  with an application to sleep-wake classification.
\newblock \emph{Frontiers in Physiology}, 12:\penalty0 637684, 2021.

\bibitem[de~Surrel et~al.(2022)de~Surrel, Hensel, Carri{\`e}re, Lacombe, Ike,
  Kurihara, Glisse, and Chazal]{de2022ripsnet}
Thibault de Surrel, Felix Hensel, Mathieu Carri{\`e}re, Th{\'e}o Lacombe,
  Yuichi Ike, Hiroaki Kurihara, Marc Glisse, and Fr{\'e}d{\'e}ric Chazal.
\newblock Ripsnet: a general architecture for fast and robust estimation of the
  persistent homology of point clouds.
\newblock In \emph{Topological, Algebraic and Geometric Learning Workshops
  2022}, pages 96--106. PMLR, 2022.

\bibitem[Edelsbrunner and Harer(2010)]{EdelsbrunnerHarer2010}
H. Edelsbrunner and J. Harer.
\newblock \emph{Computational Topology: An Introduction}.
\newblock American Mathematical Society, 2010.

\bibitem[Gumin(2016)]{gumin2016wavefunctioncollapse}
Maxim Gumin.
\newblock Wavefunctioncollapse.
\newblock \emph{GitHub repository}, 2016.

\bibitem[Hannouch and Chalup(2023)]{HannouchChalup2023}
Khalil~M. Hannouch and Stephan Chalup.
\newblock Learning to see topological properties in 4d using convolutional
  neural networks.
\newblock In \emph{Proceedings of 2nd Annual Workshop on Topology, Algebra, and
  Geometry in Machine Learning (TAG-ML), PMLR}, pages 437--454, 2023.

\bibitem[Hatcher(2002)]{Hatcher2002}
Allen Hatcher.
\newblock \emph{Algebraic Topology}.
\newblock Cambridge University Press, Cambridge, UK, 2002.

\bibitem[Krizhevsky and Hinton(2009)]{krizhevsky2009learning}
Alex Krizhevsky and Geoffrey Hinton.
\newblock Learning multiple layers of features from tiny images.
\newblock Technical Report~0, University of Toronto, Toronto, Ontario, 2009.

\bibitem[Lu et~al.(2022)Lu, Xie, Gao, Xu, and Li]{3dctn}
Dening Lu, Qian Xie, Kyle Gao, Linlin Xu, and Jonathan Li.
\newblock 3dctn: 3d convolution-transformer network for point cloud
  classification.
\newblock \emph{IEEE Transactions on Intelligent Transportation Systems}, pages
  1--12, 2022.

\bibitem[Netzer et~al.(2011)Netzer, Wang, Coates, Bissacco, Wu, and
  Ng]{netzer2011reading}
Yuval Netzer, Tao Wang, Adam Coates, Alessandro Bissacco, Bo Wu, and Andrew~Y
  Ng.
\newblock Reading digits in natural images with unsupervised feature learning.
\newblock In \emph{NIPS Workshop on Deep Learning and Unsupervised Feature
  Learning 2011}, 2011.

\bibitem[Paul and Chalup(2019)]{paul2019estimating}
Rahul Paul and Stephan Chalup.
\newblock Estimating betti numbers using deep learning.
\newblock In \emph{2019 International Joint Conference on Neural Networks
  (IJCNN)}, pages 1--7. IEEE, 2019.

\bibitem[Peek et~al.(2023)Peek, Skerritt, and Chalup]{PeekEtAl2023}
Dylan Peek, Matt Skerritt, and Stephan Chalup.
\newblock Synthetic data generation and deep learning for the topological
  analysis of 3d data.
\newblock In \emph{International Conference on Digital Image Computing:
  Techniques and Applications (DICTA 2023)}, pages 121--128. IEEE, 2023.
\newblock arXiv:2309.16968.

\bibitem[Rucco et~al.(2014)Rucco, Falsetti, Herman, Petrossian, Merelli, Nitti,
  and Salvi]{rucco2014using}
Matteo Rucco, Lorenzo Falsetti, Damir Herman, Tanya Petrossian, Emanuela
  Merelli, Cinzia Nitti, and Aldo Salvi.
\newblock Using topological data analysis for diagnosis pulmonary embolism.
\newblock \emph{arXiv preprint arXiv:1409.5020}, 2014.

\bibitem[Seifert and Threlfall(1934)]{SeifertThrelfall1934}
Herbert Seifert and William Threlfall.
\newblock \emph{Lehrbuch der Topologie}.
\newblock Teubner, Leipzig, 1934.

\bibitem[Som et~al.(2020)Som, Choi, Ramamurthy, Buman, and Turaga]{som2020pi}
Anirudh Som, Hongjun Choi, Karthikeyan~Natesan Ramamurthy, Matthew~P Buman, and
  Pavan Turaga.
\newblock Pi-net: A deep learning approach to extract topological persistence
  images.
\newblock In \emph{Proceedings of the IEEE/CVF Conference on Computer Vision
  and Pattern Recognition Workshops}, pages 834--835, 2020.

\bibitem[Wu et~al.(2015)Wu, Song, Khosla, Yu, Zhang, Tang, and Xiao]{wu20153d}
Zhirong Wu, Shuran Song, Aditya Khosla, Fisher Yu, Linguang Zhang, Xiaoou Tang,
  and Jianxiong Xiao.
\newblock 3d shapenets: A deep representation for volumetric shapes.
\newblock In \emph{Proceedings of the IEEE Conference on Computer Vision and
  Pattern Recognition}, pages 1912--1920, 2015.

\bibitem[Yamanashi et~al.(2021)Yamanashi, Kajitani, Iwata, Crutchley, Marra,
  Malicoat, Williams, Leyden, Long, Lo, et~al.]{yamanashi2021topological}
Takehiko Yamanashi, Mari Kajitani, Masaaki Iwata, Kaitlyn~J Crutchley, Pedro
  Marra, Johnny~R Malicoat, Jessica~C Williams, Lydia~R Leyden, Hailey Long,
  Duachee Lo, et~al.
\newblock Topological data analysis (tda) enhances bispectral eeg (bseeg)
  algorithm for detection of delirium.
\newblock \emph{Scientific Reports}, 11\penalty0 (1):\penalty0 1--9, 2021.

\bibitem[Yu et~al.(2021)Yu, Brakensiek, Schumacher, and Crane]{yu2021repulsive}
Chris Yu, Caleb Brakensiek, Henrik Schumacher, and Keenan Crane.
\newblock Repulsive surfaces.
\newblock \emph{arXiv preprint arXiv:2107.01664}, 2021.

\bibitem[Zhou et~al.(2022)Zhou, Dong, and Lin]{zhou2022learning}
Chi Zhou, Zhetong Dong, and Hongwei Lin.
\newblock Learning persistent homology of 3d point clouds.
\newblock \emph{Computers \& Graphics}, 102:\penalty0 269--279, 2022.

\end{thebibliography}
}

\newpage
\onecolumn
\appendix
\section{Appendix}
\subsection{Wave Function Collapse Algorithm}\label{appendix:wfc}
The following is psuedo code for a tile based implementation of the WFC algorithm. It involves the pre-determination of tiles and subsequent adjacency rules.

\begin{algorithm}
\caption{Wave Function Collapse Algorithm (Tile-Based)~\cite{gumin2016wavefunctioncollapse}}
\label{wfcalgorithm}

\SetAlgoLined
Initialize grid with uncollapsed cells\;
Initialize tile set with all possible tiles and their neighbour rules\;

\While{there are uncollapsed cells in the grid}{
    Select the cell with the lowest entropy (least number of possible tiles)\;
    \If{there are multiple cells with the same entropy}{
        Select one randomly\;
    }
    Collapse the selected cell by choosing a tile randomly from its possible tiles\;
    \tcp{Propagate constraints}
    \For{each neighbour of the collapsed cell}{
        Update the neighbour's possible tiles based on the neighbour rules\;
        \If{the neighbour's possible tiles list changes}{
            Mark the neighbour for further constraint propagation\;
        }
    }
    Propagate constraints recursively until no further changes occur\;
}

\If{grid is fully collapsed}{
    return completed grid\;
}\Else{
    handle contradiction (e.g., restart or backtrack)\;
}

\end{algorithm}

\subsection{RG Dataset Generation Time}\label{rg:generation_time}
\begin{table}[ht]
\caption{Average computation time for different genus manifolds in the `RG Repulse' dataset (minutes) using 24-core Intel Xeon Scalable ‘Cascade Lake’ processors. Samples were grown across multiple CPUs in parallel with each sequence being allocated 2 cores.} 

\centering
\begin{tabular}{l*{7}{c}}
    \toprule
    Genus 0-6 & $g_0$ & $g_1$ & $g_2$ & $g_3$ & $g_4$ & $g_5$ & $g_6$ \\
    Average Time (min) & 117.6 & 51.3 & 76.4 & 96.2 & 139.5 & 150.3 & 175.3 \\
    \addlinespace[1.05ex]
    Genus 7-13 & $g_7$ & $g_8$ & $g_9$ & $g_{10}$ & $g_{11}$ & $g_{12}$ & $g_{13}$ \\
    Average Time (min) & 192.1 & 208.1 & 209.2 & 211.1 & 226.8 & 232.1 & 227.9 \\
    \addlinespace[1.05ex]
    Genus 14-20 & $g_{14}$ & $g_{15}$ & $g_{16}$ & $g_{17}$ & $g_{18}$ & $g_{19}$ & $g_{20}$ \\
    Average Time (min) & 237.6 & 235.7 & 233.5 & 221.9 & 222.8 & 220.6 & 223.83 \\
    \bottomrule
\end{tabular}
\end{table}

\newpage
\subsection{Complexity levels} \label{appendix:complexity}
This section includes figures for objects of genus 3 and genus 10 and their corresponding complexity levels within the RG dataset.

\vspace{1em}

\noindent\begin{minipage}{\textwidth}
    \centering    
    \includegraphics[width=0.32\linewidth, trim=90 90 90 60, clip]{images/datasetRG/pointclouds/3_0_12.png}
    \includegraphics[width=0.32\linewidth, trim=90 90 90 60, clip]{images/datasetRG/pointclouds/3_1_12.png}
    \includegraphics[width=0.32\linewidth, trim=90 90 90 60, clip]{images/datasetRG/pointclouds/3_2_12.png}
    
    \vspace{0.5cm}
    
    \includegraphics[width=0.32\linewidth, trim=90 90 90 60, clip]{images/datasetRG/pointclouds/3_3_12.png}
    \includegraphics[width=0.32\linewidth, trim=90 90 90 60, clip]{images/datasetRG/pointclouds/3_4_12.png}
    \includegraphics[width=0.32\linewidth, trim=90 90 90 60, clip]{images/datasetRG/pointclouds/3_5_12.png}
    
    \vspace{0.5em}
    \captionsetup{justification=centering, font=small}
    
    \vspace{2cm}
    
    \includegraphics[width=0.32\linewidth, trim=90 90 90 60, clip]{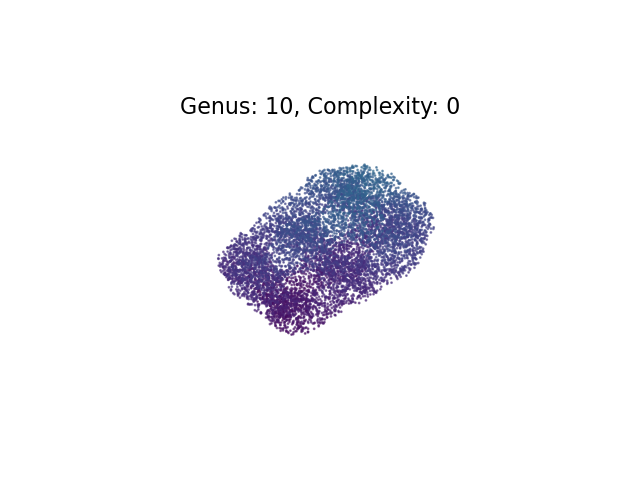}
    \includegraphics[width=0.32\linewidth, trim=90 90 90 60, clip]{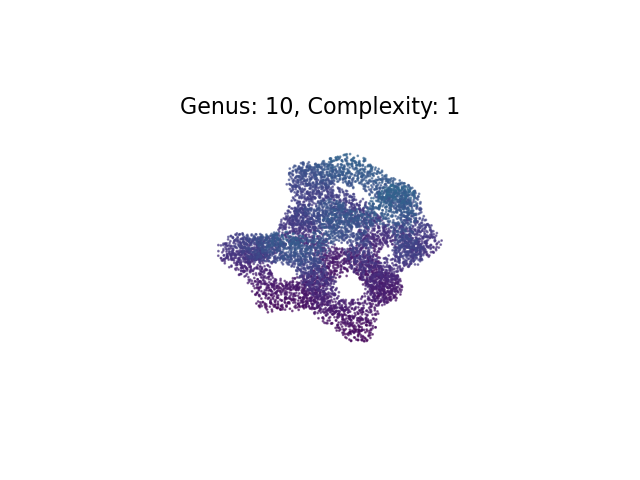}
    \includegraphics[width=0.32\linewidth, trim=90 90 90 60, clip]{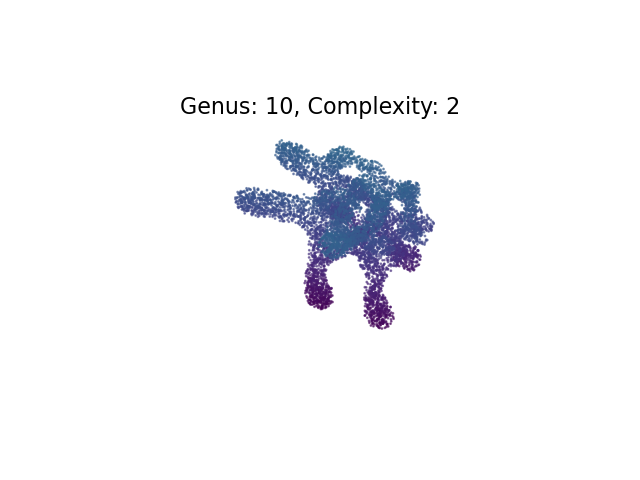}
    
    \vspace{0.5cm}
    
    \includegraphics[width=0.32\linewidth, trim=90 90 90 60, clip]{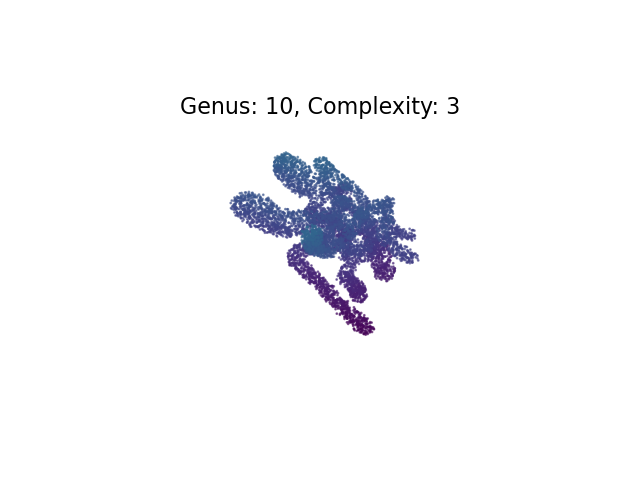}
    \includegraphics[width=0.32\linewidth, trim=90 90 90 60, clip]{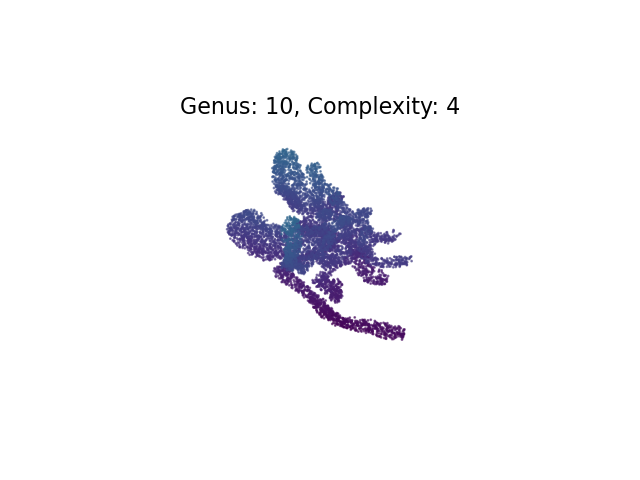}
    \includegraphics[width=0.32\linewidth, trim=90 90 90 60, clip]{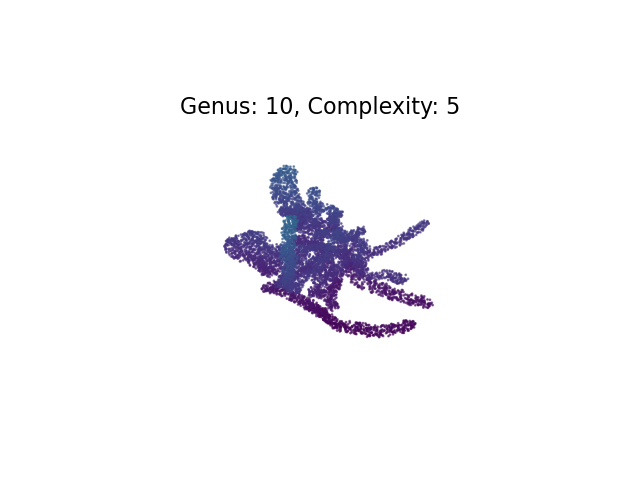}

    \captionsetup{justification=centering, font=small}
    
    \vspace{1em}
    \captionsetup{justification=justified, font=normalsize}
    \captionof{figure}{Examples of complexity levels 0--5 for genus 3 and 10 in the RG dataset. Each sample contains 8192 points.}
\end{minipage}

\newpage

\end{document}